\def\year{2021}\relax
\documentclass[letterpaper]{article} % DO NOT CHANGE THIS
\usepackage{aaai21}  % DO NOT CHANGE THIS
\usepackage{times}  % DO NOT CHANGE THIS
\usepackage{helvet} % DO NOT CHANGE THIS
\usepackage{courier}  % DO NOT CHANGE THIS
\usepackage[hyphens]{url}  % DO NOT CHANGE THIS
\usepackage{graphicx} % DO NOT CHANGE THIS
\usepackage{natbib}  % DO NOT CHANGE THIS AND DO NOT ADD ANY OPTIONS TO IT
\usepackage{caption} % DO NOT CHANGE THIS AND DO NOT ADD ANY OPTIONS TO IT
\usepackage[linesnumbered,ruled,vlined]{algorithm2e}
\usepackage{booktabs}
\usepackage{amsmath}
\usepackage{xcolor}
\usepackage{caption}
\usepackage{subcaption}

\title{CatFedAvg: Optimising Communication-efficiency and \\ Classification Accuracy in Federated Learning}
\author {
       Dipankar Sarkar,\textsuperscript{\rm 1}
       Sumit Rai, \textsuperscript{\rm 2} 
       Ankur Narang \textsuperscript{\rm 1}\\
}
\affiliations {
    \textsuperscript{\rm 1} Hike Ltd. \\
    \textsuperscript{\rm 2} IITM, Gwalior \\
    dipankars@hike.in,sumitrairkt@gmail.com,ankur@hike.in
}

\begin{document}

\maketitle

\begin{abstract}
Federated learning has allowed the training of statistical models over remote devices without the transfer of raw client data. In practice, training in heterogeneous and large networks introduce novel challenges in various aspects like network load, quality of client data, security and privacy. Recent works in FL have worked on improving communication efficiency and addressing uneven client data distribution independently, but none have provided a unified solution for both challenges. We introduce a new family of Federated Learning algorithms called CatFedAvg which not only improves the communication efficiency but improves the quality of learning using a category coverage maximization strategy. 

We  use the FedAvg framework and introduce a simple and efficient step every epoch to collect meta-data about the client's training data structure which the central server uses to request a subset of weight updates. We explore two distinct variations which allow us to further explore the tradeoffs between communication efficiency and model accuracy. Our experiments based on a vision classification task have shown that an increase of 10\% absolute points in accuracy using the MNIST dataset with 70\% absolute points lower network transfer over FedAvg. We also run similar experiments with Fashion MNIST, KMNIST-10, KMNIST-49 and EMNIST-47. Further, under extreme data imbalance experiments for both globally and individual clients, we see the model performing better than FedAvg. The ablation study further explores its behaviour under varying data and client parameter conditions showcasing the robustness of the proposed approach.
\end{abstract}

\noindent In this era of computing, we have seen an explosion of connected personal devices and IoT \cite{1}. We see tremendous progress in Deep Learning \cite{2}, which has partially been driven by the growth of data that we collect. This combination has lead to exciting research and applications in the area of distributed model training.

The conventional data processing approach has been centralized with all these devices, uploading their data into cloud servers or data centres \cite{3}. This data is then processed to provide insights or training various inference models. Recently, consumers and business alike have questioned data centralization. Governments are responding by the enactment of legislation like GDPR \cite{4} in the EU, and Consumer Privacy Bill of Rights \cite{5} in the US. These laws focus on limiting data collection and storage to what the consumer has consented for and required for processing. As data collection happens at scale, there are other technical issues like latency \cite{7} for real-time use cases and network loads.

So, we see the emergence of Mobile Edge Computing (MEC) that use the storage and processing capabilities of the edge servers and edge devices to train models closer to the data generators \cite{8}. In this solution, we transmit data to the edge servers for training a few layers, and the cloud servers do more computationally heavy tasks. This approach has shown challenges with managing communication costs and models that require continuous training \cite{9}. We see the application of Differential Privacy (DP) \cite{10} for privacy preservation, but many users are unwilling to expose their private data.  

In this regard, we see the introduction of a decentralized ML approach called Federated Learning \cite{6}. It guarantees that the data remains on the personal devices and allows collaborative machine learning of complex models. A high-level view of the approach is that we cooperatively train an ML model on a mobile device using its local data. We send the client trained model's weight to the FL server for aggregation. We repeat the steps until to reach some target accuracy. This approach is network bandwidth-efficient as only model weights are sent instead of raw data. As the model is on the mobile device, it enables privacy and makes low inference latency.

We see its successful application in production with like the Federated Averaging Algorithm (FedAvg) \cite{6} used in Google's GBoard \cite{12} to improve next-word predictions. There are other examples like recommendations \cite{13} and training of health diagnosis models which allows many hospitals and government agencies can collaborate without explicit data sharing \cite{31}. 

The Federated learning problem has a few challenges; for example, with federated networks comprising of a massive number of devices, the network data transfer costs can be humongous. There is much variability in the device capabilities in the network that leads to uncertain client behaviour. The client also generates data in a non-identically distributed manner across the network. Finally, privacy is a concern even when model weights are shared as they can reveal sensitive information about the client data. 

In this paper, we take on the dual challenge of communication efficiency and model training accuracy under client non-IID conditions. We see the use of two strategies for improving the communication efficiency, current work \cite{14} uses the static selection of a fraction of client models, and the other is the compression of the model updates \cite{15}. We further see that FedAvg performs badly when the client datasets are non-IID \cite{16} in the experiments performed. Our proposed approach improves model accuracy significantly. The contributions of the paper can be summarized as follows
\begin{itemize}
\item We propose a new family of algorithms called CatFedAvg with two proposed variants that significantly improve communication efficiency by up to 70AP while boosting model accuracy by up to 22AP. 
\item We build a framework that uses a simple and efficient client categorical coverage strategy identify the best client combinations for training. This allows us to tradeoff communication overheads for accuracy and vice-versa.
\item Our experiments show that we outperform the baseline FedAvg and VIRTUAL MTL algorithm. We solve an image classification problem using multiple image datasets under various client distribution conditions. We perform an ablation study to further show the robustness of our approach.
\end{itemize}

\section{Related Works}
\subsection{Federated Learning}
Federated learning \cite{6} is a decentralized machine learning method, which uses distributed training on clients without sending the raw data to the central server. We can split this into three categories of vertical federated learning, horizontal federated learning and federated transfer learning. 

There have been improvements proposed which, for example, improve privacy protection by the usage of differential privacy techniques \cite{10} or remove the bias of client models using agnostic federated learning \cite{22}. 

\subsection{Communication Cost}
In FL, there are several rounds of communication between the clients and the server during training. We have DL models that have millions of parameters. Thus, the larger size of the updates will lead to higher communication costs, which can be a training bottleneck. Further, unreliable network conditions and asymmetric net connection speeds will result in delays from the client. 

We can use model compression \cite{18} which is used widely in distributed learning where techniques like sparsification, quantization or subsampling \cite{19} are to reduce the model update size. However, our goal is to maintain the quality of the trained models \cite{20}. 

Alternatively, we can use selective communication where only relevant or essential updates are transmitted back \cite{21}. This technique saves communication costs and at times, improves the Global performance. We see the usage of random subsampling of clients in the sketched updating method \cite{17} to improve communication efficiency. 

\subsection{Statistical Heteroginity}

Federated learning has a few statistical challenges. In traditional distributed ML, the central server has access to the complete dataset. Allowing the server to split it into subsets with similar distributions; we transfer these to the clients for distributed training. However, this approach is impractical for FL, as the dataset is local to the client.  

In this setting, the clients may have datasets with variable distributions, i.e. the datasets are non-IID. We have cases \cite{16} where FedAvg is unable to achieve the desired accuracy in these conditions. When the dataset is highly skewed and non-IID, we have a solution where a common dataset is sent to the clients by the central server. However, this may not be possible in all cases. 

We also see the challenge of global imbalance where the collection of data held across all the clients is class imbalanced, which leads to a decrease in model accuracy. We see the Astrea framework \cite{23} trying to solve this problem, where clients initially send their data distribution to the server. Then we have a rebalancing step where each client performs data augmentation on minority classes. There is a mediator that selects the best clients that contribute to the uniform distributions. This approach has shown accuracy improvements.

There have been solutions like FEDPER \cite{24} using concepts from Multi-task learning to learn separate but structurally related models for each participant. We see the training of another set of personalization layers being trained by the client locally on its data. This has been used for building recommender systems, and testing on Flickr-AES dataset \cite{25} has shown it to outperform FedAvg. Even though federated multi-task learning has been shown to be an effective paradigm for real-world datasets, it has been applied only to convex models. We compare our algorithm performance with VIRTUAL algorithm \cite{32} with that has been designed for the federated multi-task learning with non-convex models in non-IID conditions. 

\section{CatFedAvg}

\SetKwInput{KwInput}{Input}   
\SetKwInput{KwOutput}{Output}

We introduce the category maximization federated averaging algorithm (CatFedAvg) for classification models whose primary objective is to ensure the usage of model updates which have been trained with the most number of categories. 

This is based on the FedAvg Algorithm as shown. Here \textbf{C} denotes fraction of selected clients, while \textbf{K} selected clients are indexed by \textbf{k}; \textbf{B} is the local minibatch size, \textbf{E} is the number of local epochs, and $\mathbf{\eta}$ is the learning rate.

\begin{algorithm}[t] 
\SetAlgoLined
\textbf{Server Executes:}\\
initialize $w_{0}$ \\
\For{$i \gets 1$ to $M$}{
m $\gets $ max(C.K,1)\\
$S_{t} \gets $ (random set of K clients) \label{lst:line:random_k} \\
\For{each client k $\in S_{t}$ \textbf{in parallel}}{$w_{t+1}^{k} \gets $ ClientUpdate(k,$w_{t}$)}
$w_{t+1} \gets \sum\limits_{k=1}^{k=K}\frac{n_{k}}{n}w_{t+1}$ \\}
\texttt{\\}
\textbf{ClientUpdate}(k,w): \textit{// Run on client k}\\
$\beta \gets$ (split $P_{k}$ into batches of size B)\\
\For{each local epoch i $\gets$ 1 to E}{\For{batch b $\in \beta $ }{w $\gets w - \eta \nabla l(w;b)$}}
\Return $w$ to server
\caption{\textbf{Federated Averaging Algorithm}\label{a:FedAvg}}
\end{algorithm}

Our proposal is the introduction of a meta-data collection round that makes the central system aware of what is the data distribution at the client's end. We then select a set of clients instead of the random strategy taken in FedAvg such that we maximize the coverage of categories.

We choose a classification problem that has $C$ categories, where $F$ is the total number of federated clients. In this, a random subset of $M$ is selected. This represents the larger subset whose meta-data we request. We use $N$ to denote a limiting parameter which is the maximum number of clients that we can sample. 

All the clients share their $\eta$, which is a client mask of $C$ bits, where a bit on denotes that a category is present in the new client dataset. The algorithm selects $S$ clients to maximize the coverage of $\Psi$ at the central server level. Assuming that all categories are present in the system, if we set $N = C$, we can guarantees coverage of all categories under both modes of operation.

These represent the updated client subset selection executed in line~\ref{lst:line:random_k} of the Federated Averaging Algorithm.

\begin{algorithm}[t] 
\SetAlgoLined
\KwInput{Clients Masks $\eta_{1} \dots \eta_{M}$, Global Mask $\Psi$ = 0, N, S, K=0}
\KwOutput{Set S of indices of selected Clients}
Sort $\eta_{1} \dots \eta_{M}$ in decreasing order of number of set bits.
\For{$i \gets 1$ to $C$}{
    \uIf{K = N}{\textbf{break}}
    \For{$j \gets 1$ to $M$}{
        \uIf{$\eta_{ji} = 1$ \textbf{AND} $\eta_{j} \cap S = \phi$}{
        $S \leftarrow S \cup \eta_{j}$
        $K \leftarrow K + 1$
        \textbf{break}  
        }   
    }      
}
\Return{$S$}
\caption{CatFedAvg - performance strategy}\label{a:CatFedAvg1}
\end{algorithm}

We introduce the first client selection strategy, in which select a fixed number of $C$ clients. We collect the category bitmasks, sort them in decreasing order of set bits. We then try and cover each category with one client. 

If $N \geq C$ then all categories will be covered. Further, it is possible that each category is covered more than once in the sampled model updates. This will lead to adequate data coverage for training and leads to better model accuracy.

\begin{algorithm}[t] 
\SetAlgoLined
\KwInput{Clients Masks $\eta_{1} \dots \eta_{M}$, Global Mask $\Psi$ = 0, N, S, K=0}
\KwOutput{Set S of indices of selected Clients}
Sort $\eta_{1} \dots \eta_{M}$ in decreasing order of number of set bits.

\For{$j \gets 1$ to $M$}{
    \uIf{K = N  \textbf{OR} $\Psi = 2^{C}-1$}{\textbf{break}}
    \uIf{$\Psi$ \textbf{AND} $\eta_{j} < \eta_{j}$}{
    $S \leftarrow S \cup \eta_{j}$
        
    $K \leftarrow K + 1$
    
    $\Psi \leftarrow \Psi$ \textbf{OR} $\eta_{j}$
    }
}

\Return{$S$}
\caption{CatFedAvg - cost strategy}\label{a:CatFedAvg2}
\end{algorithm}

In the cost strategy, we select a variable number of clients depending on the distribution. Similar to the first strategy, we pick a category and then select a client. However, we know that a client may have more than one category covered. That means we can skip those categories that have been covered and continue selection for the uncovered ones. We will select a maximum of C clients.  

If we have $N \geq C$ then each category is covered at least once with the least probability of having a particular category covered more than once. It optimizes cost with maximum category coverage with the goal of minimizing the number of clients required to cover all categories.

\section{Analysis}
\subsection{Federated Averaging}
We have a federated system with $M$ clients with datasets $D_{1}, D_{2}, D_{3} \dots D_{M}$ with a central server $S$. A fixed number of $K$ clients can be sampled from $M$ clients to participate in updating the global cloud model. The global loss function $F_{s}$ is related with $K$ participating clients with loss function $F_{i}$ for $i_{th}$ client as shown below.

\begin{equation}\label{eq: fs}
F_{s} = \frac{\sum\limits_{i=1}^{i=K} F_{i}}{K}
\end{equation} 

Let $F_{i}c_{j}$ denote loss contribution of $j^{th}$ class id from $i^{th}$ client. The loss contribution from $i^{th}$ client can be expressed in terms of losses of each category for a global problem with C categories, as shown below.

\begin{equation}\label{eq: fi}
F_{i} = \sum\limits_{i=1}^{i=C} F_{i}c_{j}
\end{equation}

Using eqs. (\ref{eq: fs}, \ref{eq: fi}) we have,

\begin{equation}\label{eq: fsi}
F_{s} = \frac{\sum\limits_{i=1}^{i=K}\sum\limits_{j=1}^{j=C} F_{i}c_{j}}{K}\\
KF_{s} = \sum\limits_{i=1}^{i=K}\sum\limits_{j=1}^{j=C} F_{i}c_{j}
\end{equation}

Ideally, $KF_{s}$ should have loss contribution from all categories, but in real-world skewed Non-IID conditions, some classes may not be covered by the usual random sampling technique of FedAvg technique. Hence, there may exist cases where $F_{i}c_{j}$ is not defined for any value of $i$ $\in [1, K]$

\subsection{Category coverage-based aggregation}

We define a novel method of aggregation which is focused on maximizing category coverage. Let $G_{ji}$ denote loss contribution of $j^{th}$ class from $i^{th}$ client and let $G_{j}$ denote the total loss of $j^{th}$ class id for all clients $i \in [1,K]$.

\begin{equation}\label{eq: gj}
G_{j} = \sum\limits_{j=1}^{j=C} G_{ji}
\end{equation}

We select clients in such a way that $G_{j}$ is defined for atleast one client with id $i \in [1,K]$. This is ensured using bitmasking strategy. Hence, global loss function of server $G_{s}$ can be represented as shown below.
\begin{equation}\label{eq: gs}
G_{s} = \frac{\sum\limits_{j=1}^{j=C} G_{j}}{K}
\end{equation}

Using eqs. (\ref{eq: gj}, \ref{eq: gs}) we have,
\begin{equation}
G_{s} = \frac{\sum\limits_{j=1}^{j=C}\sum\limits_{i=1}^{i=K} G_{ji}}{K}
\end{equation}

We can state that (\ref{eq: fsi}, \ref{eq: gs}) are equal. The inherent nature of averaging the loss contribution remains unchanged with our formulation.

\subsection{Understanding Tradeoffs}

\subsubsection{Communication Cost}

Let $C_{s}$ be the cost of updating the global model, assuming extra cost introduced by each client while averaging the weights to be negligible. Let $C_{c}$ be the cost incurred at the server end while interacting with a single client. The total cost for one single round with K participating clients can be approximately expressed as shown below. 

\begin{equation}\label{eq: cost}
C_{1} = KC_{c} + C_{s}
\end{equation}

The total cost for R rounds of global updates can expressed as:

\begin{gather}\label{eq: Rcost}
C_{R} = \sum\limits_{i=1}^{i=R}(KC_{c} + C_{s})\\
C_{R} = \sum\limits_{i=1}^{i=R}KC_{c} + \sum\limits_{i=1}^{i=R}C_{s}\\
C_{R} = RKC_{c} + RC{s}
\end{gather}

The cost per client interaction for R rounds can be obtained by differentiating eq (\ref{eq: Rcost}) with respect to $C_{c}$.

\begin{gather}
\frac{dC_{R}}{dC{c}} = \frac{d(RKC_{c})}{dC{c}} + \frac{d(RC_{s})}{dC{c}}\\
\frac{dC_{R}}{dC{c}} = RK + 0
\end{gather}

Dividing eq (13) by R yields cost incurred per client per round

\begin{equation}\label{eq: cost_final}
\bigg(\frac{1}{R}\bigg)\frac{dC_{R}}{dC{c}} = K
\end{equation}

Eq (14) depends directly on K. Hence, the cost incurred at the server end, increasing proportionately with the number of incorporating clients (K).

\subsubsection{Model performance}

Let the number of data samples with each of the participating client be $n_{1} \dots n_{K}$. Let the total number of data samples seen by the server be D.

\begin{equation}\label{eq: per_final}
D = \sum\limits_{i=1}^{i=K}n_{i}
\end{equation}

Hence, D increases with K. Increasing value of D help in generalizing the global model.

From eqs. (\ref{eq: cost_final},\ref{eq: per_final}) we can conclude that both cost and performance increases with number of participating clients K. 

We see that a tradeoff between high model accuracy and low cost. 

\section{Experiments}

\begin{table*}
\centering
\begin{tabular}{@{}llllllll@{}}
\hline
 &  & \multicolumn{2}{c}{\textbf{Total Samples}} & \multicolumn{4}{c}{\textbf{Samples per class}} \\ \cmidrule(lr){3-4}\cmidrule(lr){5-8}
\textbf{Dataset} & \textbf{Number of Classes} & Training & Validation & \multicolumn{2}{c}{Training} & \multicolumn{2}{c}{Validation}\\ \cmidrule(lr){5-6} \cmidrule(lr){7-8}
& & & & Mean & Std & Mean & Std \\ \midrule
\textbf{MNIST} & 10 & 60000 & 10000 & 6000 & 322 & 1000 & 59\\
\textbf{FMNIST} & 10 & 60000 & 10000 & 6000 & 0 & 1000 & 0\\
\textbf{KMNIST-10} & 10 & 60000 & 10000 & 6000 & 0 & 1000 & 0\\
\textbf{FEMNIST} & 47 & 112800 & 18800 & 2400 & 0 & 400 & 0\\
\textbf{KMNIST-49} & 49 & 232365 & 38547 & 4742 & 1839 & 786 & 309\\
\bottomrule
\end{tabular}
\caption{Description of datasets used in the experiments}
\label{table: dataset}
\end{table*}

We present a set of experiments that take compare our algorithms with FedAvg and Virtual MTL. We use five datasets and varying categories along with client data distribution. We further perform an ablation study to dissect the various parameters that show it as a clear improvement over FedAvg.

\subsection{Datasets}

We have primarily used vision datasets, and all details of the datasets used for our experiments have been summarized in Table \ref{table: dataset}.

\begin{itemize}
\item \textbf{MNIST:}  \cite{27} The dataset contains a total of 70,000 small square 28x28 pixels grayscale images of handwritten single digits between 0 and 9. The subdivision of dataset is 60,000 training samples and 10,000 testing samples. We federated the dataset amongst the clients to simulate client conditions.

\item \textbf{FMNIST} Fashion MNIST \cite{28} abbreviated as FMNIST is a dataset of Zalando's article images consisting of a training set of 60,000 examples and a test set of 10,000 examples. Each example is a 28x28 grayscale image, associated with a label from 10 classes. It is relatively challenging dataset and simulates the everyday computer vision tasks. The classes consist of fashion objects  T-shirt/top, Trouser, Pullover, Dress, Coat, Sandal, Shirt, Sneaker, Bag and Ankle Boot.

\item \textbf{KMNIST-10:} Kuzushiji-MNIST-10 (KMNIST-10) \cite{29}consists of 70,000 28x28 pixels grayscale images of handwritten Japanese characters of Hiragana. The subdivision of this dataset is 60,000 training samples and 10,000 testing samples. It is a well-balanced dataset across all classes and is relatively difficult to learn as compared to MNIST. We use this to simulate the federated settings.

\item \textbf{FEMNIST:} The  FEMNIST  dataset is the federated version of EMNIST dataset \cite{30}. It is a set of handwritten character digits derived from the NIST Special Database 19 and converted to a 28x28 pixel image format and dataset structure that directly matches the MNIST dataset.  We used the balanced version with ten single-digit classes between 0 and 9 inclusive, 26 uppercase alphabets and 11 lowercase alphabets. We federated this 47 classes dataset for the simulation of real-world federated settings.

\item \textbf{KMNIST 49:} Kuzushiji-MNIST 49 is a much larger, but imbalanced dataset containing 48 Hiragana characters and one Hiragana iteration mark.  It is an excellent example to simulate the universal problem of an imbalanced class problem under federated settings. It consists of 49 classes 28x28 grayscale images. It is a comparatively much larger dataset.

\end{itemize}

\subsection{Data distributions}

\begin{figure*}
\begin{subfigure}{.2\textwidth}
  \centering
  \includegraphics[width=3cm, height=4cm, keepaspectratio]{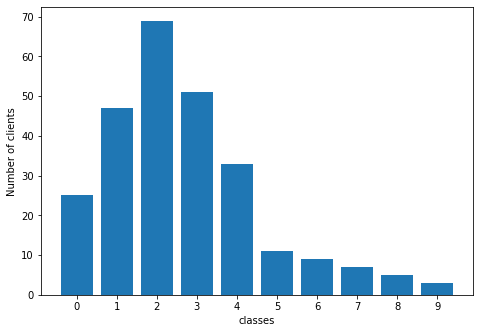}
  \caption{D$_{1}$}
  \label{fig:sfig1}
\end{subfigure}%
\begin{subfigure}{.2\textwidth}
  \centering
  \includegraphics[width=3cm, height=4cm, keepaspectratio]{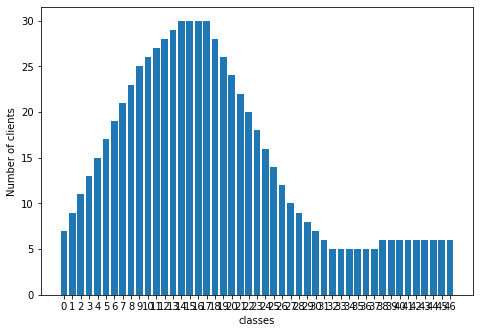}
  \caption{D$_{2}$}
  \label{fig:sfig2}
\end{subfigure}
\begin{subfigure}{.2\textwidth}
  \centering
  \includegraphics[width=3cm, height=4cm, keepaspectratio]{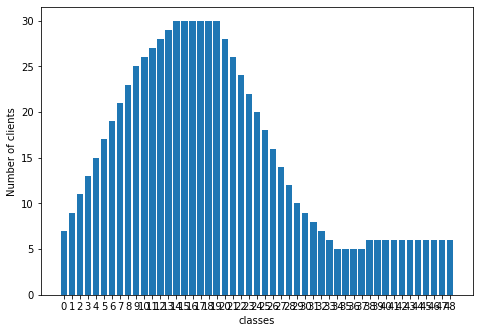}
  \caption{D$_{3}$}
  \label{fig:sfig3}
\end{subfigure}
\begin{subfigure}{.2\textwidth}
  \centering
  \includegraphics[width=3cm, height=4cm, keepaspectratio]{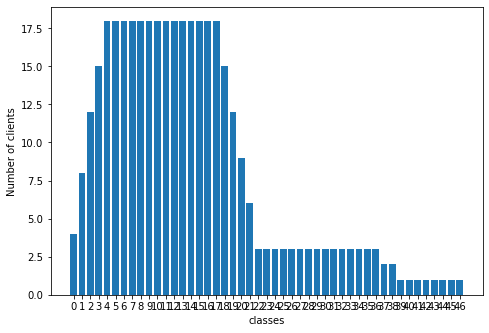}
  \caption{D$_{4}$}
  \label{fig:sfig4}
\end{subfigure}
\begin{subfigure}{.16\textwidth}
  \centering
  \includegraphics[width=3cm, height=4cm, keepaspectratio]{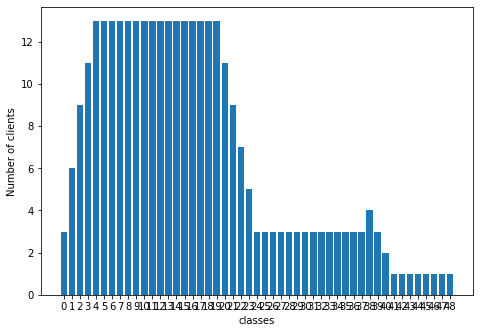}
  \caption{D$_{5}$}
  \label{fig:sfig5}
\end{subfigure}

\caption{Distribution of Categories across clients (Category ids on x-axis and number of clients on y-axis)}
\label{fig: distribution_categories}
\end{figure*}

We simulate various client data distribution conditions to compare the algorithm. We take a total of 100 clients in the Federated system. We vary the number of clients having a particular category. In all of our experiments, all clients have a proper subset of categories varying from clients having a single category to clients having half the number of total categories. There are five distribution conditions (D$_{1}$, D$_{2}$, D$_{3}$, D$_{4}$, D$_{5}$) which we use and have been shown in figure \ref{fig: distribution_categories}. We share specific details in the Appendix.

\subsection{Settings}

\begin{figure}[!h]
\centering
\includegraphics[scale=0.5]{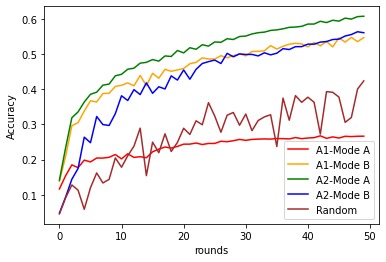}
\caption{Performance comparison of proposed algorithms on distribution D$_{5}$}
\label{fig: comparison_d5}
\end{figure}

All networks employed are multilayer perceptrons (MLP) with two hidden dense Flipout layers with 100 units in case of MNIST, single hidden layer of 512 units in case of FMNIST and KMNIST-10 and a single hidden layer of 784 units in case of FEMNIST-47 and KMNIST-49. Activation function Rectified Linear Unit (ReLU) is employed between the hidden layers and input layers, while the softmax activation function is employed in the final output layer. The final output layer consists of 10 units in case of MNIST, FMNIST and KMNIST-10. In FEMNIST and KMNIST-49 number of final layer units are 47 and 49 respectively. 

We define two modes of experiments per algorithm where we set the value of the maximum number of clients to sample. We define  Mode A where N is kept at ten and Mode B, where we keep N as equal to the number of categories in the dataset. This allows us to explore conditions where we may want to sample as few clients as possible. We will denote the algorithms for performance strategy to be A1 and cost strategy to be A2. 

Each client has 600 training samples, and batch size is 32 is used in the execution of subroutine of local refinements for all experiments. In the case of a random client selection method, which is the default FedAvg case, a total of 10 clients are selected for participation in global training. 

For all cases, the learning rate is used as 0.003, and a total of 50 rounds of global refinements are performed using Stochastic Gradient Descent (SGD) with Federated Averaging (FedAvg) method of updating the global model.

\subsection{Results}

\begin{table*}[t]
 \centering
 \begin{tabular}{@{}lllllll@{}}
     \midrule
 \textbf{Dataset} & \textbf{FedAvg} & \textbf{Mode A - A1} & \textbf{Mode A - A2} & \textbf{Mode B - A1} & \textbf{Mode B - A2} & \textbf{Virtual MTL} \\
 \midrule
MNIST & 0.7285$\pm$0.0092 & 0.9202$\pm$0.0007 & 0.8343$\pm$0.0080 & \textbf{0.9445$\pm$0.004} & 0.8371$\pm$0.0060 & 0.5711$\pm$0.0060 \\
 FMNIST & 0.5622$\pm$0.0230 & 0.7937$\pm$0.0020 & 0.7084$\pm$0.0050 & \textbf{0.8087$\pm$0.003} & 0.6829$\pm$0.0010 & 0.4432$\pm$0.0060 \\
 KMNIST-10 & 0.5776$\pm$0.0070 & 0.8071$\pm$0.0004 & 0.7586$\pm$0.0030 & \textbf{0.8129$\pm$0.007} & 0.7498$\pm$0.0040 & 0.4721$\pm$0.0060\\
 \bottomrule
 \end{tabular}
 \caption{Accuracy of global model on test set using random and proposed algorithms on distribution D$_{1}$}
 \label{table: result1}
\end{table*}

\begin{table*}[t]
 \centering
 \begin{tabular}{@{}lllllll@{}}
     \midrule
 \textbf{Dataset} & \textbf{Distribution}  & \textbf{FedAvg} & \textbf{Mode A - A1} & \textbf{Mode A - A2} & \textbf{Mode B - A1} & \textbf{Mode B - A2} \\
 \midrule
  FEMNIST & D2 & 0.5573$\pm$0.0130 & 0.4123$\pm$0.0005 & 0.6903$\pm$0.006 & \textbf{0.7537$\pm$0.002} & 0.6921$\pm$0.002 \\
  & D4 & 0.4123$\pm$0.023 & 0.2712$\pm$0.0007 & 0.5818$\pm$0.008 & \textbf{0.7023$\pm$0.001} & 0.6446$\pm$0.002 \\
 KMNIST-49 & D3 & 0.5211$\pm$0.0150 & 0.3972$\pm$0.0050 & 0.6213$\pm$0.004 & \textbf{0.6753$\pm$0.001} & 0.6286$\pm$0.001 \\
  & D5 & 0.4236$\pm$0.0270 & 0.2653$\pm$0.0080 & 0.5325$\pm$0.008 & \textbf{0.6157$\pm$0.005} & 0.5795$\pm$0.005 \\

 \bottomrule
 \end{tabular}
 \caption{Accuracy of global model on test set using random and proposed algorithms on KMNIST-49 and FEMNIST}
 \label{table: result2}
\end{table*}

\begin{table*}
\centering
\begin{tabular}{@{}lllllllllllll@{}}
\hline
\textbf{Distribution/(Classes)} & \multicolumn{2}{c}{\textbf{Mode A-A1}} & \multicolumn{2}{c}{\textbf{Mode A-A2}} & \multicolumn{2}{c}{\textbf{Mode B-A1}} & \multicolumn{2}{c}{\textbf{Mode B-A2}} \\ \cmidrule(lr){2-3} \cmidrule{4-5} \cmidrule{6-7} \cmidrule{8-9} &  \textbf{Selected} & \textbf{Covered} & \textbf{Selected} & \textbf{Covered} & \textbf{Selected} & \textbf{Covered} & \textbf{Selected} & \textbf{Covered}\\ 
\midrule
\textbf{D$_{1}$ (10)}  & 10 & \textcolor{blue}{10} & 3 & \textcolor{blue}{10} & \textcolor{red}{10} & \textcolor{blue}{10} &3 & \textcolor{blue}{10}  \\
\textbf{D$_{2}$ (47)}  & 10 & 24 & 6 & \textcolor{blue}{47} & \textcolor{red}{47} & \textcolor{blue}{47} &6 & \textcolor{blue}{47}  \\
\textbf{D$_{3}$ (49)}  & 10 & 24 & 6 & \textcolor{blue}{49} & \textcolor{red}{49} & \textcolor{blue}{49} &6 & \textcolor{blue}{49} \\
\textbf{D$_{4}$ (47)}  & 10 & 14 & 10 & 39 & \textcolor{red}{47} & \textcolor{blue}{47} &19 & \textcolor{blue}{47} \\
\textbf{D$_{5}$ (49)}  & 10 & 14 & 10 & 39 & \textcolor{red}{49} & \textcolor{blue}{49} &19 & \textcolor{blue}{49}\\
\bottomrule
\end{tabular}
\caption{Description of clients selected (\textbf{Selected}) in each round and categories covered (\textbf{Covered}) using the proposed algorithms on various distributions. ( Highest number of clients selected is marked \textcolor{red}{red} and all categories covered is marked in \textcolor{blue}{blue} )}
\label{table: distribution_actual}
\end{table*}

Here we describe the various insights about the experimental results as presented in Tables \ref{table: result1} and \ref{table: result2}.

We discuss the number of clients selected, as that reflects the communication efficiency along with the model accuracy numbers under various conditions. We further performed an ablation study which has been shared in the Appendix.

\subsubsection{FedAvg}
We pick a subset of 10 clients is selected randomly for global refinement. It performs well when the distribution of classes is not skewed. However, for the skewed environments of D$_{1}$  with ten categories, we see a dip in performance as depicted in Table \ref{table: result1}. The deterioration is pronounced when a large number of categories (47 and 49) are there as seen in distributions D$_{2}$ and D$_{3}$. The primary reason is the lack of data coverage of all categories in the simulated skewed environment is not guaranteed. The dataset conditions D$_{4}$ and D$_{5}$ show the worst performance.

\subsubsection{CatFedAvg - performance strategy} 

Here we primarily discuss the algorithm A1 which implements the model performance strategy.\\
\textbf{Mode A} : We limit the total number of clients selected to $N=10$. In this scenario, all ten categories were covered in case of MNIST, FMNIST and KMNIST-10 under distribution D${1}$. In case of distribution D$_{2}$, and D$_{3}$ using FEMNIST and KMNIST-49 only 24 categories out of 47 and 49 respectively were covered while in D$_{4}$ and D$_{5}$ only 14 categories were covered as described in \ref{table: distribution_actual}. This results in poor performance of this strategy as the category coverage is not enough. 

\textbf{Mode B}: We don't limit the number of clients selected in this mode. Hence, coverage of all categories is guaranteed by this algorithm. We see that clients are specifically selected for each category which leads to the selection of 10 clients in distribution D$_{1}$, 47 clients in distributions D$_{2}$ and D$_{4}$ and 49 clients in distributions D$_{3}$ and D$_{5}$. This performs the best in comparison to other algorithms as evident from Tables \ref{table: result1} and \ref{table: result2}.

\subsubsection{CatFedAvg - cost strategy}
Here we primarily discuss the algorithm A2 which implements the communication efficiency strategy.\\
\textbf{Mode A}:  We limit the total number of clients selected to $N=10$. It is seen that only three clients are selected in case of distribution D$_{1}$ while six clients are selected in case of D$_{2}$ and D$_{3}$. In D$_{4}$ and D$, _{5}$ total clients, selected are ten which is the maximum limit (N=10). Hence the coverage of all categories is not guaranteed here, and only 39 categories were covered in each round, but it performs better than the performance strategy variant in the same mode. 

\textbf{Mode B}: We don't limit the number of clients selected in this mode. All categories are covered with minimal probability of a category being selected more than once. In this case, three clients are selected in distribution D$_{1}$ while six clients are selected in distributions D$_{2}$ and D$_{3}$. In the extremely skewed distributions D$_{4}$ and D$_{5}$, selected of 19 clients is sufficient for complete coverage.

\section{Observations}

\subsection{Model Accuracy}
As described in \ref{table: result1} and \ref{table: result2}, the performance of Algorithm 1 under mode B is the best across the entire spectrum under all conditions. 

If we denote the number of categories in global problem begin solved as C, the relative performance of proposed algorithms is as below.

If N$\geq$C: 
\textbf{Mode A-A1 $\approx$ Mode A-A2 $\approx$ Mode B-A1 $\approx$ Mode B-A2}

If N$<$C: 
\textbf{Mode B-A1 $\geq$ Mode B-A2 $\geq$ Mode A-A2 $\geq$ Mode A-A1}

\subsection{Communication Cost}

Communication cost is related to the efficiency of a particular algorithm when it comes to deployment, and the cost of refining the global model increases with the number of clients participating in the training round. \ref{table: distribution_actual} describes the number of clients selected in various distributions.

Considering all distributions, $D_{1} \dots D_{5}$, we can see that number of clients participating is least in case of  Mode A-A2. We limit the clients participating in mode A-A1 to N. In contrast; Mode B-A2 has a comparatively higher number of participating clients when the number of global categories is less than N in mode A-A2.

Let C be the number of global categories. The following is the relative order of algorithms in terms of communication cost expense.
 
If N$\geq$C: 
\textbf{Mode A-A1 $\approx$ Mode B-A1 $\geq$ Mode A-A2 $\approx$ Mode B-A2}

If N$<$C: 
\textbf{Mode B-A1 $\geq$ Mode B-A2 $\geq$ Mode A-A1 $\geq$ Mode A-A2}

\subsection{Trade Off : Accuracy versus Communication}

We can see that the better performance of mode B-A1 comes at a higher communication cost. If we want the best performance, we need to compromise with communication cost. Alternatively, if we desire a communication efficiency, then Mode A-A2 is desirable. A better compromise is achieved with both performance and communication cost may suggest Mode B-A2 as the best one. 

\section{Conclusions \& Future Work}

In this work, we have proposed an approach that can help tradeoff communication overheads with model accuracy with the usage of a simple and efficient technique. This family of algorithm CatFedAvg has been shown to perform better by 10AP in some cases than existing well-proven methods like FedAvg and VIRTUAL MTL. The communication efficiency has seen a jump of 70AP in some cases. We show a formal analysis that gives a framework to understand how to work with the critical parameters for the tradeoff between communication efficiency and model accuracy.

We have also shown the robustness of the approach with various statistical heterogeneous conditions like global class imbalance and client distribution variations. We believe that existing techniques that involve data augmentation may not be feasible in production conditions. We would see this approach combined with compression to deliver even more communication efficiency. It would be interesting to use latency to be factored in when the initial larger set of random clients are selected. We offer a technique that can possibly work easily with Differential Privacy due to its inherent simplicity offering privacy as well. 

\bibliography{index.bib}
\end{document}

\maketitle
\section{Data distribution}

\begin{table}[t]
 \centering
 \begin{tabular}{@{}lll@{}}
     \midrule
 \textbf{Distribution} &\textbf{Global classes} & \textbf{Categories per client}\\
 \midrule
 D$_{1}$ & 10 & 1-5  \\
 D$_{2}$ & 47 & 1-15   \\
 D$_{3}$ & 49 & 1-15  \\
 D$_{4}$ & 47 & 1-5  \\
 D$_{5}$ & 49 & 1-5  \\

 \bottomrule
 \end{tabular}
 \caption{Description of range of number of categories on each client in the distribution }
\label{table: distribution}
\end{table}

These are the various distributions that we have used to test the algorithm's performance. 

\begin{itemize}
\item \textbf{D$_{1}$ :} Datasets having ten categories ( MNIST, FMNIST and KMNIST-10 ) are used to distribute categories across all clients according to this distribution. Presence of a particular category on the number of clients varies from 3-70 with each client having a variable number of categories ranging from single category to $50\%$ of total categories. Lower category ids are denser while higher category ids are present on very few clients as described in Figure X (a).

\item \textbf{D$_{2}$ :} The 47 classes of FEMNIST dataset are distributed according to this distribution with the presence of a particular category on several clients varying from 6-30 with each client having a variable number of categories ranging from single category to 30\% of total categories. Intermediate category ids are densely populated among the clients as compared to other classes.

\item \textbf{D$_{3}$ :} KMNIST-49 comprising of the maximum number of classes is distributed according to this distribution where category distribution forms a deformed bell-shaped curve with intermediate category ids being densely populated. The higher is lower category ids are least in number with the presence of particular category on several clients varying from 6-30 with each client having a variable number of categories ranging from single category to 30\% of total categories.

\item \textbf{D$_{4}$ :} This distribution is a variant of D$_{2}$. It is a more skewed version where a presence of a particular category on several clients varied from 1 to 13 with each client having a variable number of categories ranging from single category to 10\% of total categories.

\item \textbf{D$_{5}$ :} This distribution is another extremely skewed variation of D$_{3}$ where nearly half the number of categories is very scarce along with the presence of single category clients. Presence of a particular category varies from 1 to 18 with each client having a variable number of categories ranging from single category to 10\% of total categories. in case. 
\end{itemize}  

\section{Ablation Study}

We perform an ablation study to describe the insights of the importance of category-based client selection criteria. Firstly, the effect of varying the number of categories on clients is described, and then the variation of the upper limit parameter of Algorithm 1 is described in detail. 

\subsection{Variation of number of categories}

\begin{table}[t]
 \centering
 \begin{tabular}{@{}llllll@{}}
     \midrule
 \textbf{Dataset} & \textbf{D$_{6}$} & \textbf{D$_{7}$} & \textbf{D$_{8}$} & \textbf{D$_{9}$} & \textbf{D$_{10}$} \\
 \midrule
 MNIST & 1 & 3 & 5 & 7 & 9  \\
 FMNIST & 1 & 3 & 5 & 7 & 9  \\
 KMNIST-10 & 1 & 3 & 5 & 7 & 9  \\
 FEMNIST & 1 & 3 & 10 & 25 & 35  \\
 KMNIST-49 & 1 & 3 & 10 & 25 & 35  \\

 \bottomrule
 \end{tabular}
 \caption{Distribution of number of categories on each client for different datasets for ablation study}
 \label{table: ablation_study}
\end{table}

\begin{figure*}
\begin{subfigure}{.3\textwidth}

  \includegraphics[width=5cm, height=6cm, keepaspectratio]{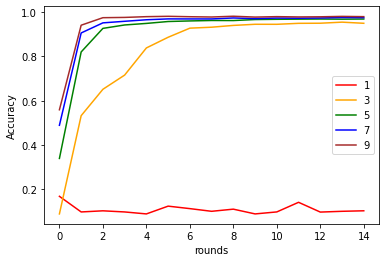}
  \caption{MNIST}
  \label{fig:sfig1}
\end{subfigure}%
\begin{subfigure}{.3\textwidth}

  \includegraphics[width=5cm, height=6cm, keepaspectratio]{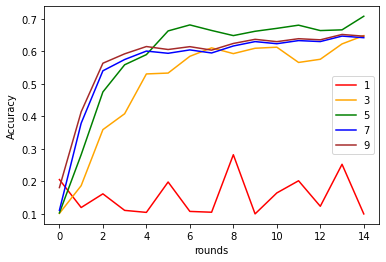}
  \caption{FMNIST}
  \label{fig:sfig2}
\end{subfigure}
\begin{subfigure}{.3\textwidth}

  \includegraphics[width=5cm, height=6cm, keepaspectratio]{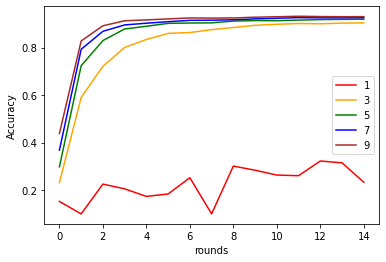}
  \caption{KMNIST-10}
  \label{fig:sfig3}
\end{subfigure}\\
\hspace*{2.7cm}
\begin{subfigure}{.3\textwidth}

  \includegraphics[width=5cm, height=6cm, keepaspectratio]{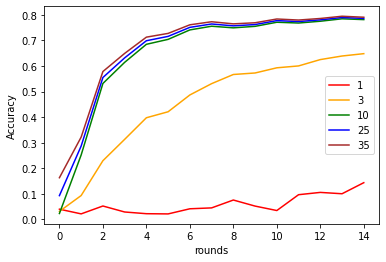}
  \caption{FEMNIST}
  \label{fig:sfig4}
\end{subfigure}
\begin{subfigure}{.3\textwidth}

  \includegraphics[width=5cm, height=6cm, keepaspectratio]{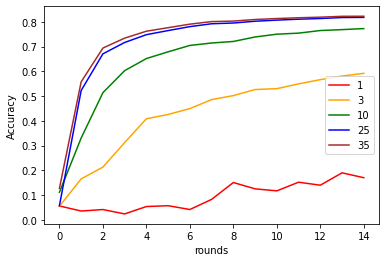}
  \caption{KMNIST-49}
  \label{fig:sfig5}
\end{subfigure}

\caption{Variation of number of categories under different distributions.}
\label{fig: variation_category}
\end{figure*}

\begin{table*}[t]
 \centering
 \begin{tabular}{@{}llllll@{}}
     \midrule
\textbf{Dataset} & \textbf{FedAvg} & \textbf{Mode A - A1} & \textbf{Mode A - A2} & \textbf{Mode B - A1} & \textbf{Mode B - A2} \\
 \midrule
 MNIST & 0.6594$\pm$0.0092 & 0.8585$\pm$0.0007 & 0.5623$\pm$0.0080 & \textbf{0.8949$\pm$0.004} & 0.3423$\pm$0.0060 \\
 FMNIST & 0.4566$\pm$0.0230 & 0.7066$\pm$0.0020 & 0.6312$\pm$0.0050 & \textbf{0.7461$\pm$0.003} & 0.6521$\pm$0.0010  \\
 KMNIST-10 & 0.5234$\pm$0.0070 & 0.7431$\pm$0.0004 & 0.7353$\pm$0.0030 & \textbf{0.7576$\pm$0.007} & 0.7328$\pm$0.0040 \\
 \bottomrule
 \end{tabular}
 \caption{Imbalanced Case :Accuracy of global model on test set using random and proposed algorithms on distribution D$_{1}$}
 \label{table: global_imbalance}
\end{table*}

This study focused on how changing the number of categories on each client affect the performance of training global model under federated settings. For this purpose, we simulate five distributions ( D$_{6}$ \dots D$_{10}$ ) for each of the five datasets considered. The nature of distributions ( D$_{6}$ \dots D$_{10}$ ) for each of the dataset is described in Table \ref{table: ablation_study}. The graphical representation of training for each of the mentioned distribution is presented in Figure \ref{fig: variation_category}. 

We note that when the system consists of clients with the only single category, the global model is not able to perform. As the number of categories is increased, the performance of global model keeps on improving with distribution D$_{10}$ of all datasets performing the best. 

\subsection{Variation of parameter N}

\begin{figure}[!h]
\centering
\includegraphics[scale=0.5]{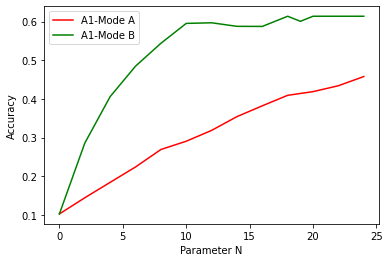}
\caption{Impact of variation of parameter N in Mode A-A1 and Mode A-A2 on distribution D$_{5}$}
\label{fig: ablation_variation}
\end{figure}

We study the variation of the client limiting parameter N for both algorithms. We use the rather interesting extreme distribution D$_{5}$ of KMNIST-49 dataset to study the variation in performance.

In Figure \ref{fig: ablation_variation}, we can see that as the value of limiting parameter is varied from 1 to 25, the performance using algorithm one under mode A and mode B vary differently. As the value of N is increased, the performance of mode B improves much faster than mode A. This is because mode B covers more categories with the same value of N as compared to mode A. 

The optimal value of N found is 19 to cover all categories under this distribution. It can be fruitful to increase N while using mode A to cover all categories and achieve comparable performance.

\section{Global Distribution Imbalance}

We investigated the performance of FedAvg and CatFedAvg under class imbalance in distribution D1. Table  \ref{table: global_imbalance}  shows the results on three datasets MNIST, FMNIST and KMNIST-10 for ten categories out of which four categories have very few samples in the ratio of 1:10. 

We found that CatFedAvg outperforms FedAvg by using the method of category maximization. Although in both cases, there is no focus on imbalance a simple maximization of category coverage helps in better performance.